\pdfoutput=1

\documentclass[11pt]{article}

\usepackage[final]{acl}

\usepackage{times}
\usepackage{latexsym}
\usepackage{amsmath}
\newtheorem{definition}{Definition}[section]

\usepackage{amssymb}
\usepackage[T1]{fontenc}

\usepackage[utf8]{inputenc}

\usepackage{microtype}
\usepackage{booktabs}
\usepackage{multirow}

\usepackage{inconsolata}

\usepackage{graphicx}

\title{Optimizing Multi-Hop Document Retrieval Through Intermediate Representations}

\author{
 \textbf{Jiaen Lin\textsuperscript{1{$\star$}}},
 \textbf{Jingyu Liu\textsuperscript{3{$\star$}}},
 \textbf{Yingbo Liu\textsuperscript{1, 2}}
\\
 \textsuperscript{1}School of Software, BNRist, Tsinghua University,
\\
 \textsuperscript{2}Beijing Key Laboratory of Industrial Bigdata System and Application,
 \\
 \textsuperscript{3}Gaoling School of Artificial Intelligence, Renmin University of China,
\\
 \small{
   \textbf{Correspondence:} \href{mailto:lje23@mails.tsinghua.edu.cn}{lje23@mails.tsinghua.edu.cn}
 }
}

\usepackage{amsmath,amsfonts,bm}

\def\eqref#1{equation~\ref{#1}}

\def\1{\bm{1}}

\def\vh{{\bm{h}}}

\def\vl{{\bm{l}}}

\def\vp{{\bm{p}}}

\def\vs{{\bm{s}}}

\def\vu{{\bm{u}}}
\def\vv{{\bm{v}}}

\def\vx{{\bm{x}}}

\def\mK{{\bm{K}}}

\def\mQ{{\bm{Q}}}

\def\mU{{\bm{U}}}
\def\mV{{\bm{V}}}
\def\mW{{\bm{W}}}
\def\mX{{\bm{X}}}

\def\mSigma{{\bm{\Sigma}}}

\DeclareMathAlphabet{\mathsfit}{\encodingdefault}{\sfdefault}{m}{sl}
\SetMathAlphabet{\mathsfit}{bold}{\encodingdefault}{\sfdefault}{bx}{n}

\def\tdwv{{\mathrm{TD}(\mW_v)}}

\newcommand{\Ls}{\mathcal{L}}

\newcommand{\softmax}{\mathrm{softmax}}

\begin{document}
\maketitle
\let\thefootnote\relax\footnotetext{$^\star$ Equal Contribution\hspace{3pt}}
\begin{abstract}
Retrieval-augmented generation (RAG) encounters challenges when addressing complex queries, particularly multi-hop questions. While several methods tackle multi-hop queries by iteratively generating internal queries and retrieving external documents, these approaches are computationally expensive.
In this paper, we identify a three-stage information processing pattern in LLMs during layer-by-layer reasoning, consisting of extraction, processing, and subsequent extraction steps. This observation suggests that the representations in intermediate layers contain richer information compared to those in other layers.
Building on this insight, we propose Layer-wise RAG (L-RAG). Unlike prior methods that focus on generating new internal queries, L-RAG leverages intermediate representations from the middle layers, which capture next-hop information, to retrieve external knowledge.
L-RAG achieves performance comparable to multi-step approaches while maintaining inference overhead similar to that of standard RAG.
Experimental results show that L-RAG outperforms existing RAG methods on open-domain multi-hop question-answering datasets, including MuSiQue, HotpotQA, and 2WikiMultiHopQA. The code is available in \url{https://github.com/Olive-2019/L-RAG}.

\end{abstract}

\section{Introduction}

Large Language Models (LLMs) have made significant advancements in natural language processing, demonstrating exceptional performance across a variety of tasks \cite{muennighoff-etal-2023-mteb, hendryckstest2021}, including complex multi-hop tasks. However, despite their success, LLMs often generate factually incorrect answers as they rely solely on their internal parametric memory, which struggles with tasks requiring external knowledge \cite{RAG}. To address this limitation, Retrieval-Augmented Generation (RAG) frameworks \cite{gao2023retrieval, liu2024much} have been developed, utilizing semantic similarity metrics to retrieve relevant external documents, which are subsequently integrated into the generation context. Nevertheless, RAG still faces challenges, particularly in handling complex queries that require multi-hop document retrieval.

\begin{figure}[t]
  \includegraphics[width=\columnwidth]{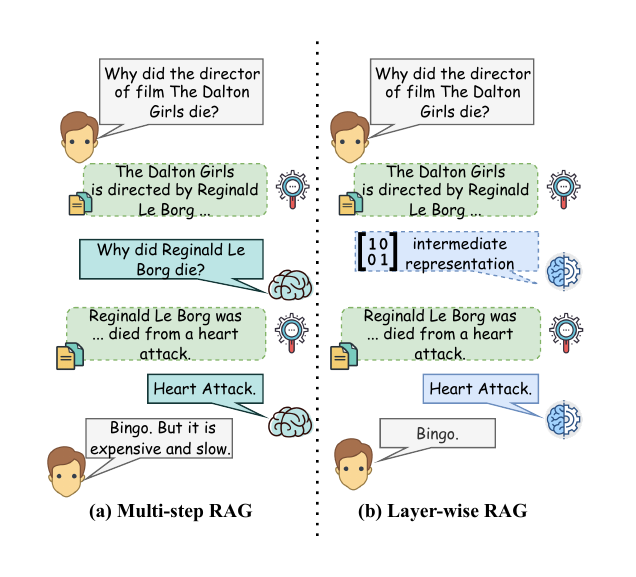}
  \caption{Comparison of Multi-step RAG and Layer-wise RAG. The left panel illustrates Multi-step RAG, which involves multiple rounds of reasoning to generate retrieval queries. In contrast, the right panel shows Layer-wise RAG, which requires only a single round of reasoning to produce intermediate representations for retrieval.}
  \label{fig:process}
\end{figure}
Multi-step methods, such as Self-ask \cite{self-ask} and ITER-RETGEN \cite{iter-retgen}, address the challenges of multi-hop reasoning and document retrieval by decomposing complex queries into manageable steps and iteratively refining the retrieval process \cite{active-RAG, enhancing-RAG, query-rewriting-in-RAG}. 
Over multiple iterations, the response is refined until a final answer is produced. 
As illustrated in Figure \ref{fig:process}, while multi-step methods significantly improve retrieval and reasoning capabilities, they often result in higher computational overhead due to their iterative nature. Several variants aim to address this issue. Adaptive-RAG \cite{adaptive-rag}  simplify the reasoning process for less complex queries. However, for more complex queries, multiple full inference iterations are still required. Efficient RAG \cite{efficient-rag} get rid of relying on LLM calls at each step, using the smaller model to generate the special token for information processing. Despite these advancements, such methods still tend to address the problem in generating human understandable query, overlooking the next-hop information contained in intermediate representations during reasoning. There is a question we try to find out that \textbf{how can we leverage the information contain to achieve performance comparable to multi-step approaches while maintaining a low inference overhead similar to vanilla RAG?}

LLMs demonstrate the ability to extract entities and their relationships during inference \cite{factual-association}, including implicit reasoning for bridge entities in multi-hop tasks \cite{hopping-too-late, implicit-reasoners, mechanistic-analysis-on-multi-step-reasoning-task}. Moreover, studies suggest that intermediate representations within LLMs often contain richer information than the final layer \cite{intermediate-representation}. Through an analysis of LLM weight matrices using Singular Value Decomposition (SVD), we identify the computational regime of LLMs: initial layers primarily perform raw information extraction, middle layers engage in contextual processing, and final layers focus on extracting information for token generation. This analysis suggests that intermediate layer representations, rather than those in the final layer, are more suitable for document retrieval.

In light of this, we introduce \textbf{Layer-wise Retrieval-Augmented Generation (L-RAG)}, which leverages intermediate representations from the middle layers to retrieve external information. As illustrated in Figure \ref{fig:overview-of-LRAG}, we train a contriever-based representations retriever to effectively utilize these intermediate representations for retrieving higher-hop documents.

Our experimental results demonstrate that the proposed method achieves competitive performance while maintaining computational efficiency comparable to vanilla-RAG systems.

Our key contributions include: 
\begin{itemize}
    \item We propose an SVD-based decomposition approach to analyze weight matrices, identifying the three-stage information processing pattern in LLMs, and validate this through the LogitLens method. 
    \item We introduce a novel method that utilizes intermediate representations to retrieve external information, which is then incorporated into the prompt for final generation. 
    \item We rigorously evaluate the proposed method across several open-source LLMs, demonstrating its effectiveness in enhancing inference performance. 
\end{itemize}
\section{Related Works}

\subsection{Retrieval-Augmented Generation}
\begin{figure*}
    \centering
    \includegraphics[width=\linewidth]{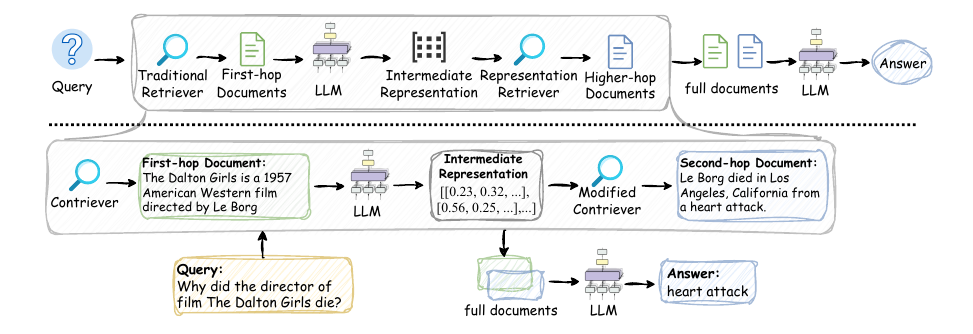}
    \caption{An overview and example of the L-RAG framework, which consists of three components: a traditional retriever, an LLM for generating intermediate representations, and the representation retriever. The area above the dashed line represents the framework, while the area below illustrates the demonstration. Initially, the traditional retriever (e.g., BM25, Contriever) retrieves relevant chunks from the corpus as the first-hop document. The LLM (e.g., LLaMA2-7B) then uses the query and first-hop context to generate an intermediate representation. This intermediate representation is used by the Modified Contriever trained by us to retrieve the higher-hop document. Finally, the LLM generates the final answer using the complete context.}
    \label{fig:overview-of-LRAG}
\end{figure*}

To address factual inaccuracies in language models, Retrieval-Augmented Generation (RAG) integrates external knowledge by providing relevant passages as contextual input, thereby enhancing factual reasoning and enabling real-time utilization of information. However, RAG systems often struggle with complex, multi-hop queries \cite{hotpotqa, MuSiQue, 2wikimultihop}, which require iterative synthesis of information. 

Recent advancements in RAG have introduced multi-round retrieval strategies, including query rewriting \cite{query-rewriting-in-RAG, enhancing-RAG, active-RAG} and self-questioning \cite{self-asking}, aimed at improving performance on complex tasks. Nevertheless, these methods face a significant limitation: they incur increased computational latency and cost due to the repeated invocation of large language models (LLMs) for query generation. To address these challenges, several methodological variants have been proposed. Adaptive-RAG \cite{adaptive-rag} reduces inefficiencies by dynamically assessing query complexity and selectively activating retrieval processes, thus lowering inference overhead for simpler queries while retaining multi-round retrieval for more complex cases. Alternatively, Efficient RAG \cite{efficient-rag} uses a smaller model to iteratively generate queries and filter irrelevant information, bypassing the need for repeated LLM calls, thereby improving computational efficiency.

\subsection{Reasoning Mechanisms of Transformers}

A lines of research has employed empirical approaches to explore the internal workings of transformers \cite{neural-networks-chomsky-hierarchy, structural-recursion, layersllmsnecessaryinference} and their reasoning mechanisms \cite{recall-idioms, factual-association}, particularly in the context of multi-step reasoning tasks \cite{mechanistic-analysis-on-multi-step-reasoning-task, hopping-too-late, mechanistic-interpretation-of-arithmetic-reasoning}.

Our work is motivated by the findings of \citet{intermediate-representation}, who found that intermediate representations often retain more comprehensive information compared to final layer representations. Drawing from this observation, we utilize Singular Value Decomposition (SVD) to build a novel analytical framework, shifting the focus from hidden representations to the analysis of weight matrices. This approach offers a more generalized and interpretable perspective for understanding model behavior.
\begin{figure*}[t]
  \includegraphics[width=\linewidth]{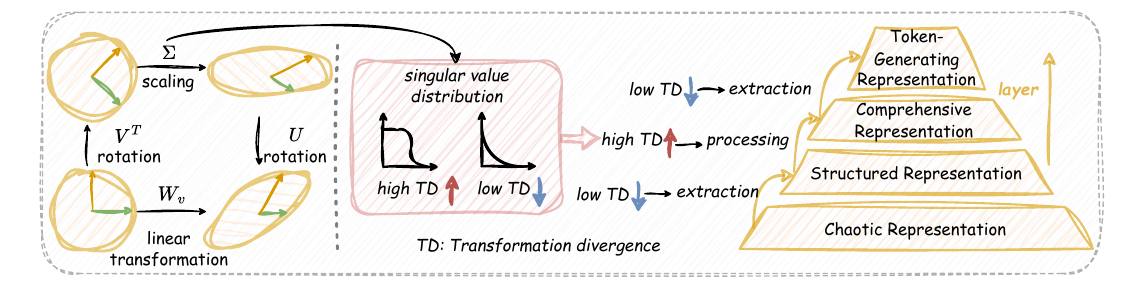}
  \caption{Weight matrix transformation divergence (TD) and information processing in LLM reasoning. The left panel illustrates the SVD, where the weight matrix is decomposed into three operators, including scaling. The singular value distribution represents the scaling magnitude of the weight matrix, which is quantified using TD. The right panel shows how the value of TD reflects the information processing stages in LLM reasoning.}
  \label{fig:svd-and-llm-reasoning}
\end{figure*}
\section{Analysis of LLM Reasoning Mechanisms}
We systematically analyze how LLMs process information across layers, revealing distinct reasoning phases through singular value decomposition and empirical validation.

\subsection{Transformer-based Language Models}
Transformer-based language models consist of multiple transformer blocks, each comprising an attention layer (\textit{Attn}) and a Multi-Layer Perceptron (MLP), connected via a residual stream \cite{transformers}.

The attention layer utilizes multiple attention heads, each performing scaled dot-product attention:
\begin{equation*}
   \text{Attn}(\mX) = \softmax \left( \frac{\mW_q \vx \cdot (\mW_k \vx)^T}{\sqrt{d_k}}\right ) \mW_v \vx ,
\end{equation*}
where $d_k$ represents the dimensionality of the key vectors, $\vx$ denotes the input vector, and $\mW_q$, $\mW_k$, and $\mW_v$ are the weight matrices for the queries, keys, and values, respectively. The softmax operation generates an attention weight distribution over the input sequence, which indicates the relevance of each token to the current position. The value $\mV = \mW_v \vx$, encapsulating the informational content, is weighted by these attention probabilities, enabling the dynamic aggregation of contextual information across the sequence. Consequently, the output of the attention block is essentially a weighted sum of $\mV$, with the weights computed based on the interaction between $\mQ$ and $\mK$. Therefore, $\mW_v$ plays a pivotal role in facilitating the flow of information within the attention mechanism, and analyzing it provides valuable insights into the processing and propagation of information within the framework. This analysis is further explored using Singular Value Decomposition (SVD).

\subsection{Transformation Divergence}
\label{sec:transformaiton-divergence}
The linear transformation $\mW_v \vx$ can be interpreted as a combination of rotation and scaling operations applied to the vector $\vx$, and these operations are elegantly captured through Singular Value Decomposition (SVD). SVD decomposes a matrix into three simpler matrices, providing a geometric interpretation of the transformation represented by the original matrix.

For a matrix $\mW_v$ of size $m \times n$, SVD expresses it as:
\begin{equation}
\begin{split}
    \mW_v &= \mU \mSigma \mV^T \\
          &=  \vu_1 \sigma_1 \vv_1^T + \dots + \vu_r \sigma_r \vv_r^T \\
          &= \sigma_1 \vu_1 \vv_1^T + \dots + \sigma_r \vu_r \vv_r^T,
\end{split}
\label{eq:svd}
\end{equation}

where $r$ denotes the rank of $\mW_v$, and $\mSigma = \text{diag}(\sigma_1, \sigma_2, \dots, \sigma_r)$ is a diagonal matrix of size $m \times n$ containing the singular values of $\mW_v$, which are non-negative and sorted in descending order, while matrices $\mU = [\vu_1, \dots, \vu_m]$ and $\mV = [\vv_1, \dots, \vv_n]$ are orthogonal matrices of sizes $m \times m$ and $n \times n$, respectively, with $\vu_i$ and $\vv_i$ representing the left and right singular vectors.

This decomposition provides a geometric interpretation of the transformation represented by $\mW_v$. The matrix $\mV$ applies a rotation in the input space, while $\mSigma$ scales the input vectors by the singular values. Finally, $\mU$ performs a rotation in the output space, as illustrated in Figure \ref{fig:svd-and-llm-reasoning}. In essence, the linear transformation can be decomposed into two components: rotation, which maps vectors to other space, and scaling, which stretches or shrinks the roation operations. The transformation directions are defined by the matrices $\mU$ and $\mV$.

\begin{definition}[Transformation Direction]
    \label{transformation-direction}
    The transformation direction of a transformer matrix $\mW_v$ is formally defined as:
    \begin{equation*}
        \mathrm{Direction}(\mW_v) = [\vu_1 \vv_1^T, \dots, \vu_r \vv_r^T]
    \end{equation*}
\end{definition}

The linear transformation $\mW_v \vx$ can be expressed in terms of these transformation directions, since $\mW_v \vx = \sigma_1 \vu_1 \vv_1^T \vx + \dots + \sigma_r \vu_r \vv_r^T \vx$ according to Equation \ref{eq:svd}, where $\sigma$ is the scaling value that emphasizes the magnitude of the transformation along each direction. 

Scaling operations are mathematically characterized by the singular values $\sigma$, which represent the relative magnitudes of the orthogonal transformation directions in the matrix decomposition. When the singular values are similar, it indicates that the transformation processes information across multiple directions. In contrast, large discrepancies between singular values suggest a more selective extraction of information. Therefore, singular values can be used to determine whether the transformation focuses on information extraction or processes information across different directions.

\begin{definition}[Transformation Divergence]

The transformation divergence of any non-zero matrix (transformation) $\mW_v \in \mathbb{R}^{m \times n}$ is defined as:
\begin{equation}
    \mathrm{TD}(\mW_v) = - \sum_{i=1}^{r} \frac{\sigma_i}{\sum_{j=1}^{r} \sigma_j} \log \frac{\sigma_i}{\sum_{j=1}^{r} \sigma_j} ,
\end{equation}
where $r = \min(m, n)$, and $\sigma_1, \sigma_2, \dots, \sigma_r$ are the singular values of matrix $\mW_v$.
\label{eq:transformation-divergence}
\end{definition}

Transformation divergence, as defined above, quantifies the extent of information processing and extraction within a linear transformation represented by a matrix. It focuses on relative magnitudes rather than absolute ones, specifically measuring the disparity in the scaling of different directions. A higher transformation divergence, such as for the vector $(5,4,3)^T$, with a divergence value of 1.08, indicates a transformation that processes information across multiple directions. In this case, the non-zero values suggest that all directions will be retained, though some directions are weighted more heavily than others. 
In contrast, a lower transformation divergence signifies a more concentrated focus on a particular direction. For instance, for the vector $(1,0,0)^T$, the transformation divergence indicates that the transformation primarily extracts information from a single dominant direction, discarding information from other directions. This metric allows for a layer-wise analysis of the model's behavior, as illustrated in Figure \ref{fig:svd-and-llm-reasoning}.

\subsection{Layer-wise Dynamics of Transformation Divergence}
\label{sec:preliminary-test}
To show how different layers work in LLM using transformation divergence, we applied Equation \ref{eq:transformation-divergence} to quantify this metric across diverse open-source LLMs with varying architectures. These models exhibit fundamental structural differences, rather than differing in parametric scale, as detailed in Appendix \ref{sec:appendix:transformation-divergence}.

\begin{figure}
    \centering
    \includegraphics[width=\linewidth]{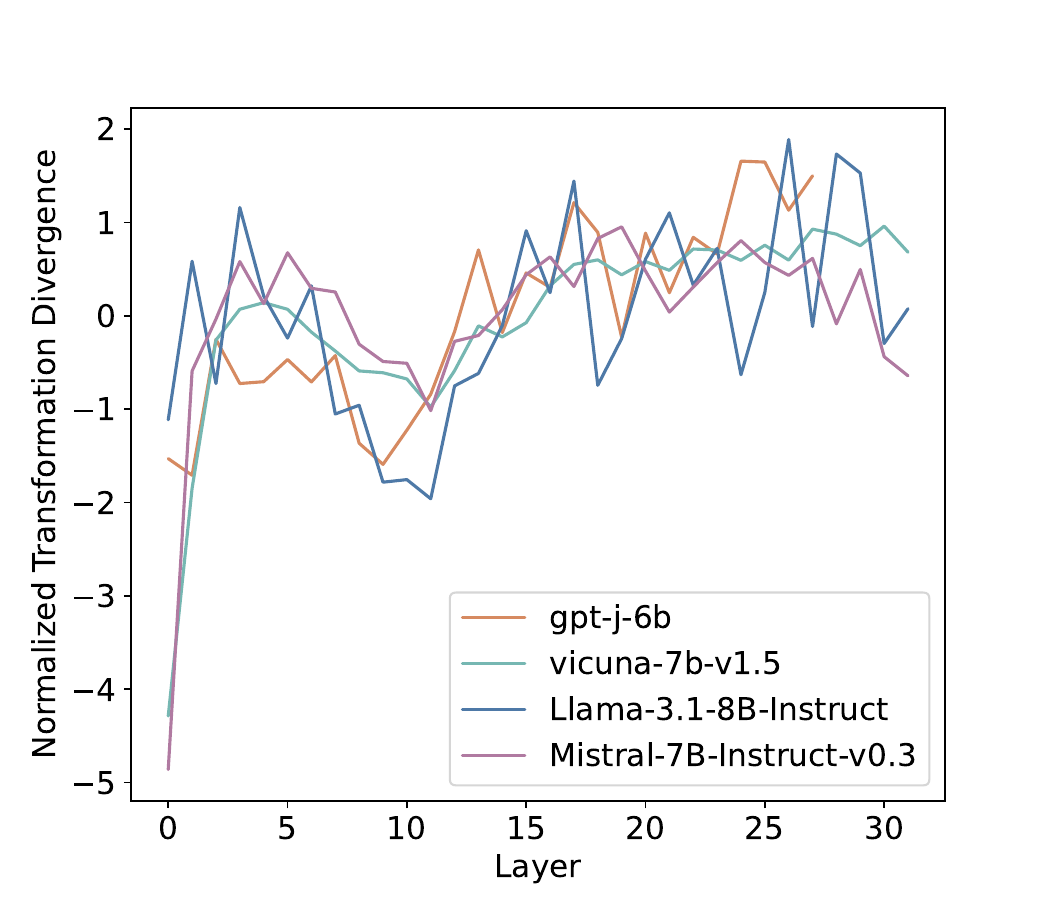}
    \caption{Transformation Divergence $\tdwv$ of the weight matrix across various models.}
    \label{fig:transformation-divergence}
\end{figure}

As illustrated in Figure \ref{fig:transformation-divergence}, transformation divergence $\tdwv$ exhibits a consistent three-phase progression across all analyzed models: an initial high-value phase, followed by a period of progressive decrease, and a final increase. As discussed in Section \ref{sec:transformaiton-divergence}, low $\tdwv$ corresponds to a concentrated extraction of information along principal component directions, while high $\tdwv$ indicates broad manipulation across multiple transformation directions. Based on this, we hypothesize that LLMs undergo three distinct reasoning operations, leading to four representational states.

These three operations reflect fundamental information processing mechanisms within transformer architectures. During the initial low $\tdwv$ phase, LLMs extract information from disorganized data. The subsequent high $\tdwv$ phase signifies the processing of this information. Finally, the resurgent low $\tdwv$ phase aligns the representation with the token prediction requirements which is also proved in \citet{layersllmsnecessaryinference} and \citet{recall-idioms}. Based on these operations, we can categorize the representations during inference into four types:

\begin{itemize}
    \item \textbf{Chaotic Representation}: The initial representation containing raw input features.
    \item \textbf{Structured Representation}: The structured information extracted from the chaotic representation.
    \item \textbf{Comprehensive Representation}: A multi-dimensional embedding that integrates structured information, contextual dependencies, and domain-specific knowledge for a holistic understanding.
    \item \textbf{Token-Generating Representation}: The final output specialization aligning for token prediction.
\end{itemize}

Consequently, the structured and comprehensive representations may be particularly effective for retrieving relevant next-hop documents.

To validate this hypothesis, we evaluate the relationship between transformation divergence and information processing in LLMs by computing the recall rate for documents indirectly related to the query using different layer representations in the L-RAG framework, as described in Section \ref{sec:method}. These results are then compared with the transformation divergence patterns. As shown in Appendix \ref{sec:td-recall-rate}, the best recall performance is observed at the layer during or after the low transformation divergence period, supporting the hypothesis outlined above.

\subsection{LogitLens Across LLMs}

To validate the transformation divergence pattern, we apply the LogitLens technique to track the evolution of information across transformer layers. Our hypothesis suggests that, during layer-by-layer reasoning, LLMs extract and process information progressively. This process likely generates intermediate information, which would result in intermediate answers emerging earlier than the final answers.

We analyze individual hidden states using the LogitLens technique \cite{recall-idioms}, where activations are transformed into logits for each vocabulary token by multiplying them with the output embedding matrix, as detailed in Appendix \ref{sec:appendix:logit-lens}.

\begin{figure}
    \centering
    \includegraphics[width=\linewidth]{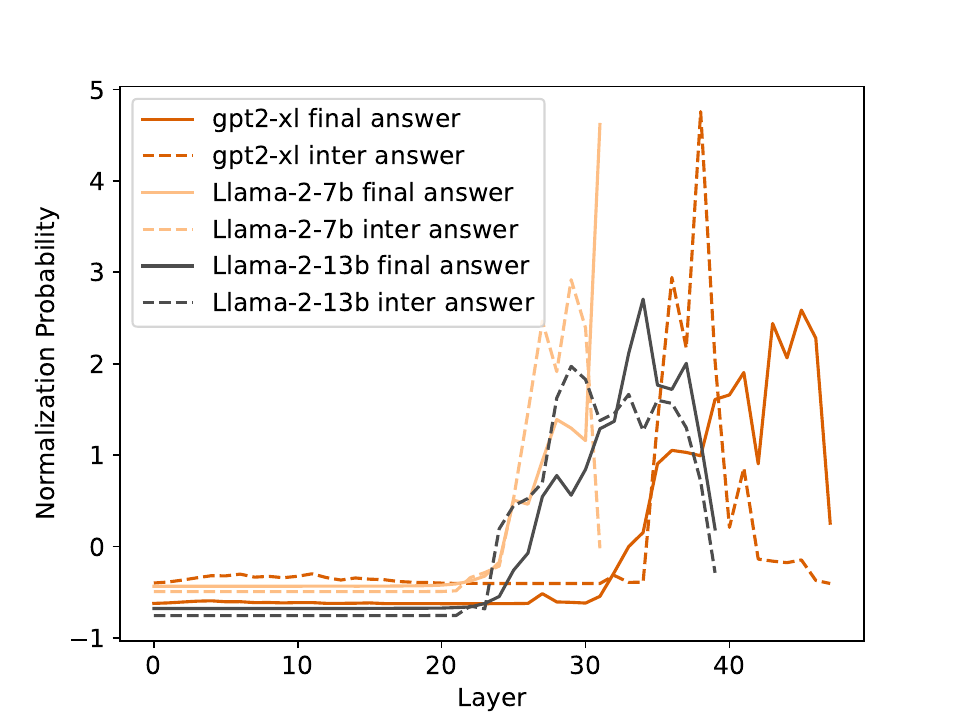}
    \caption{Logit Lens analysis conducted at multiple layers within different LLMs. The dashed lines represent the probability of intermediate answers, while the solid lines indicate the probability of final answers.}
    \label{fig:logit-lens}
\end{figure}

As shown in Figure \ref{fig:logit-lens}, intermediate answers emerge earlier than final answers across various LLMs. This suggests that the information leading to the intermediate answer is processed earlier in the reasoning process. Therefore, identifying the layer with the most relevant information for the intermediate answer is crucial for retrieving the next-hop document, rather than relying on the final layer.

\section{Method}
\label{sec:method}

We present Layer-wise Retrieval-Augmented Generation (L-RAG), an innovative framework that enhances retrieval-augmented LLMs by leveraging intermediate representations to improve the efficiency of multi-hop document retrieval, particularly at higher hops. In Section \ref{sec:preliminary-test}, we discuss how LLMs extract valuable information from chaotic representations, process this information into comprehensive representations, and ultimately derive token-generation information for producing token-generating representations. Based on this understanding, we proposed an overview of the L-RAG framework as shown in Figure \ref{fig:overview-of-LRAG}. This section outlines the pipeline structure in Section \ref{sec:overview-of-lrag}, followed by the detailed implementation procedures in Section \ref{sec:training}.

\subsection{Overview of L-RAG}
\label{sec:overview-of-lrag}

L-RAG consists of three main components. Initially, the traditional retriever (such as BM25 or Contriever) retrieves the first-hop documents directly related to the query. These documents are then combined with the query to form an input instruction. LLM perform single token-level reasoning to generate an intermediate representation. Representation retriever, designed to effectively utilize the nuanced information embedded in these intermediate representations, then retrieves next-hop documents that are more contextually relevant.

Figure \ref{fig:overview-of-LRAG} illustrates the L-RAG framework, detailing the sequential steps from query input to final answer generation. By integrating these components, L-RAG aims to improve the overall retrieval efficiency and answer quality in retrieval-augmented LLMs.

\subsection{Implementation Details}
\label{sec:training}

In the L-RAG framework, only the representation retriever requires training. We utilize off-the-shelf traditional retrievers and LLMs. For representation retriever, we add a MLP layer in front of the original Contriever to align LLM intermediate representations with the Contriever embedding space. The relevance score between an intermediate representation and a document is calculated as the dot product of their respective representations after processing through the Contriever.
Formally, given an intermediate representation $r$ and a document $d$, both are encoded independently using the same model $f$. Additionally, the intermediate representation $r$ is passed through an MLP $g$. The relevance score $s(r, d)$ is then defined as:

\begin{equation}
    s(r, d) = f(g(r))^T \cdot f(d)
    \label{equation:similarity}
\end{equation}

For training, we employ contrastive InfoNCE loss to enable the model to distinguish between documents by comparing positive pairs (related documents) and negative pairs (unrelated documents) using intermediate representations. Formally, given a batch of intermediate representations $R = [r_1, r_2, \dots, r_n]$ and a pool of associated positive documents $D = [d_1, d_2, \dots, d_m]$, the contrastive loss for one representation and its corresponding positive document is defined as:
\begin{equation}
    \hat\Ls(r_i, d_i^+) = \frac{\exp(s(r_i, d^+) / \tau)}{\sum_{d \in D} \exp(s(r_i, d) / \tau)},
    \label{equation:single-loss}
\end{equation}

where $s$ represents the similarity score, as defined in Equation \ref{equation:similarity}, and $\tau$ is the temperature parameter.

The overall InfoNCE loss for the set of intermediate representations $R$ is computed as the average loss across all representations $r \in R$. It is formally defined as:
\begin{equation}
    \Ls(R, D) =- \log\sum_{r_i \in R} \sum_{d^+_i}\hat\Ls(r_i, d^+_i), 
\end{equation}

where $d_i^+$ denotes the positive document for representation $r_i$ in $D$, and $\hat\Ls(r_i, d^+_i)$ is the contrastive InfoNCE loss defined in Equation \ref{equation:single-loss}. We train L-RAG separately for each dataset to achieve better performance.

For reasoning, the layer with the minimum TD is a natural starting point for selection. L-RAG would refine the choice by exploring neighboring layers to identify the most suitable one. Specifically, we define the set of candidate layers (CL) as:
\begin{equation}
    CL(k, n, l) = l - nk, \dots, l, \dots, l + nk,
\end{equation}
where $n$ represents the number of candidate layers (CL) considered on each side, and $k$ denotes the step size. We then train retrievers using each layer in candidate layers and evaluate their performance. The layer yielding the best results is selected as the optimal one.

\section{Experiment}
\begin{table}[htbp]
  \centering
    \begin{tabular}{cccc}
    \toprule
    \textbf{Method} & \textbf{2Wiki} & \textbf{Musique} & \textbf{Hotpot} \\
    \midrule
    No Retrieval & 2.29 & 2.30 & 2.28 \\  
        VanillaRAG@4 & 2.60 & 2.55 & 2.64 \\  
        VanillaRAG@8 & 2.79 & 2.67 & 2.73 \\  
        SelfAsk & 14.80 & 19.62 & 19.59 \\  
        IterRetGen & 20.07 & 21.18 & 42.36 \\  
        IRCoT & 25.47 & 14.57 & 17.30 \\  
        HyDE & 5.60 & 4.81 & 5.28 \\  
        L-RAG@4 & 2.76 & 2.62 & 2.80 \\  
        L-RAG@8 & 2.87 & 2.86 & 2.92 \\  
    \bottomrule
    \end{tabular}%
  \caption{Task latency per query across different datasets for the Llama3.1-8B model.}
  \label{tab:latency}%
\end{table}%

\begin{figure}
    \centering
    \includegraphics[width=\linewidth]{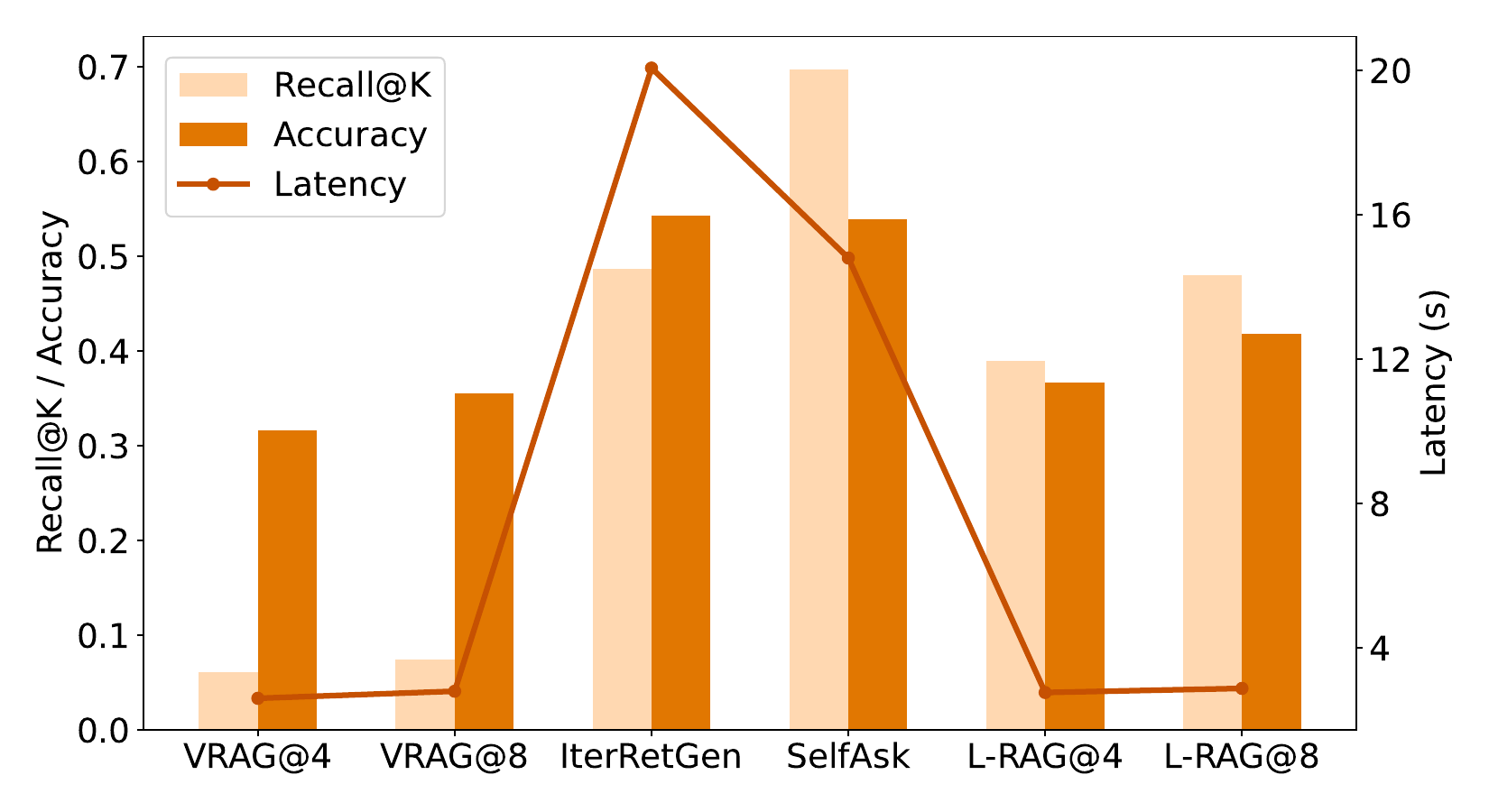}
    \caption{The recall rate and accuracy of different methods in the 2WikiMQA dataset. VRAG stands for Vanilla RAG.}
    \label{fig:hit-rate-lentency}
\end{figure}
\subsection{Setup}

\textbf{Datasets.} We evaluate our method on three multi-hop question answering (QA) benchmark, which require sequential reasoning across multiple documents: MuSiQue \cite{MuSiQue} under CC-BY 4.0 license, HotpotQA \cite{hotpotqa} under a CC BY-SA 4.0 license, and 2WikiMultiHopQA \cite{2wikimultihop} under Apache license 2.0 . Specifically, we focus on queries that require two-hop reasoning to arrive at the correct answer, explicitly excluding those solvable through single-hop inference, as detailed in Appendix \ref{sec:appendix:multi-hop-dataset-example}. Additionally, we use GPT-4 to identify higher-hop documents that cannot be directly recognized within the dataset.

\textbf{Baselines.} Our baseline models encompass: (1) straightforward methods, specifically \textbf{No Retrieval}, \textbf{Vanilla RAG}, and \textbf{HyDE} \cite{HyDE}; (2) advanced multi-step retrieval strategies, namely \textbf{SelfAsk} \cite{self-ask}, \textbf{IterRetGen} \cite{iter-retgen} and \textbf{IRCoT} \cite{IRCoT}. All models are executed on the LLama3.1-8b platform.

\textbf{Metrics.} To evaluate L-RAG, we assess three aspects: retrieval performance, task performance, and task efficiency. For retrieval performance, we report the higher-hop document recall rate. Task performance is measured by accuracy (Acc), in line with standard evaluation protocols. Accuracy is determined by whether the predicted answer contains the ground-truth answer. Regarding efficiency, we report the average time required to answer each query.

\subsection{Task Efficiency}

We calculate the latency of each query during the execution of those method.
As shown in Table \ref{tab:latency} and Figure \ref{fig:hit-rate-lentency}, our method requires low latency, comparable to Vanilla-RAG and direct reasoning, while delivering significantly better performance.

\subsection{Retrieval Performance}
\begin{table*}[htbp]
  \centering
    \begin{tabular}{ccccccc}
    \toprule
          & \multicolumn{2}{c}{\textbf{2WikiMQA}} & \multicolumn{2}{c}{\textbf{Musique}} & \multicolumn{2}{c}{\textbf{HotpotQA}} \\
    \textbf{Method} & \textbf{Recall@K} & \textbf{K} & \textbf{Recall@K} & \textbf{K} & \textbf{Recall@K} & \textbf{K} \\
    \midrule
    Vanilla RAG@4 & 0.061 & 4.0 & 0.695 & 4.0 & 0.543 & 4.0 \\ 
        Vanilla RAG@8 & 0.074 & 8.0 & 0.762 & 8.0 & 0.641 & 8.0 \\ 
        IterRetGen & 0.487 & 6.0 & 0.523 & 6.0 & \textbf{0.732} & 6.0 \\ 
        SelfAsk & \textbf{0.697} & 27.4 & 0.765 & 28.1 & 0.711 & 28.6 \\ 
        IRCoT & 0.665 & 11.1 & \textbf{0.830} & 14.1 & 0.770 & 7.9 \\ 
        HyDE & 0.071 & 6.0 & 0.280 & 6.0 & 0.250 & 6.0 \\ 
        L-RAG@4 & 0.389 & 4.0 & 0.520 & 4.0 & 0.449 & 4.0 \\ 
        L-RAG@8 & 0.480 & 8.0 & 0.764 & 8.0 & 0.613 & 8.0 \\
    \bottomrule
    \end{tabular}%
  \caption{Results on higher-hop document recall rate across three dataset. Baselines are implemented from the source code. Bold fonts denote the best results in the same generator.}
  \label{tab:retrieval-performance}%
\end{table*}%

\label{sec:retrieval-performance}

We evaluate the model's performance on next-hop document retrieval using the Recall@K metric across three datasets. As shown in Table \ref{tab:retrieval-performance}, L-RAG demonstrates moderate yet acceptable recall scores. Notably, it significantly outperforms Vanilla RAG on the 2WikiMQA dataset and achieves near-best performance on the Musique dataset.

\begin{figure}
    \centering
    \includegraphics[width=\linewidth]{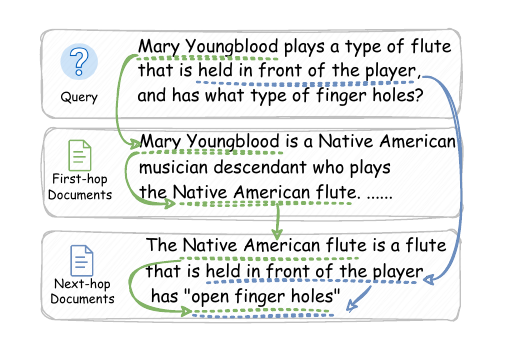}
    \caption{Information Leakage in Multi-Hop Reasoning on HotpotQA. This diagram contrasts standard reasoning pathways (green arrows) with shortcut retrieval patterns (blue arrows), demonstrating how query leverage surface-level positional descriptors (e.g., 'held in front of the player') to directly match target documents while bypassing critical intermediate reasoning about cultural attributes of Native American flutes.}
    \label{fig:information-leakage}
\end{figure}

The recall metric for the vanilla RAG method quantifies information leakage in higher-hop document retrieval, reflecting the likelihood of retrieving higher-hop documents using the initial query.
As shown in Table \ref{tab:retrieval-performance}, 2WikiMQA dataset exhibits the lowest direct recall score (0.061), while other datasets achieve recall rates above 0.5. This significant gap suggests that 2WikiMQA contains less inherent information leakage between reasoning hops compared to the other benchmarks. As Figure \ref{fig:information-leakage} shown, while the question structure ostensibly requires intermediate reasoning about the "Native American flute" concept, the presence of the distinctive positional descriptor "held in front of the player" creates an unintended retrieval shortcut.
Therefore, 2WikiMQA serves as a more rigorous and reliable benchmark for evaluating multi-hop reasoning, as its structure reduces shortcut solutions through direct query matching, requiring genuine multi-step inference.
\subsection{Task Performance}
\label{sec:task-performance}

\begin{table}[htbp]
  \centering
    \begin{tabular}{cccc}
    \toprule
    \textbf{Dataset} & \textbf{2Wiki} & \textbf{Musique} & \textbf{Hotpot} \\
    \midrule
    No Retrieval & 0.275 & 0.094 & 0.230 \\
    VanillaRAG@4 & 0.316 & 0.305 & 0.590 \\
    VanillaRAG@8 & 0.355 & 0.297 & 0.598 \\
    SelfAsk & 0.539 & 0.180 & 0.373 \\
    IterRetGen & \textbf{0.543} & 0.297 & \textbf{0.668} \\
    IRCoT & 0.498 & 0.346 & 0.615 \\ 
    HyDE & 0.174 & 0.076 & 0.248 \\ 
    L-RAG@4 & 0.367 & 0.258 & 0.555 \\
    L-RAG@8 & 0.418 & \textbf{0.352} & 0.641 \\
    \bottomrule
    \end{tabular}%

  \caption{Task performance (Acc) across different method and three dataset.Bold fonts denote the best results in the same generator.}
  \label{tab:task-performance}%
\end{table}%

As shown in Table \ref{tab:task-performance}, the L-RAG framework demonstrates superior performance across benchmark datasets, achieving the highest accuracy (0.352) on MuSiQue, securing a competitive second-place ranking (0.641) on HotpotQA, and attaining third place (0.418) on the complex 2WikiMQA corpus. While L-RAG may not achieve the best performance across all datasets, its task latency, as illustrated in Figure \ref{fig:hit-rate-lentency}, indicates that its performance is sufficiently strong for practical real-world applications.
\section{Conclusion}
This work has three principal contributions to enhance multi-hop reasoning in Retrieval-Augmented Generation (RAG) systems. First, we identify a three-stage computational pattern in LLM inference through singular value decomposition (SVD) of weight matrices and LogitLens analysis. Second, building on this insight, we propose Layer-wise RAG (L-RAG), a novel framework that leverages intermediate layer representations for document retrieval. By training a Contriever-based retriever to interpret these representations, our method bypasses iterative LLM calls while capturing critical multi-hop reasoning signals. Third, comprehensive evaluations across multiple LLMs demonstrate that L-RAG achieves competitive accuracy with multi-step RAG methods, yet maintains computational efficiency comparable to vanilla RAG. Our findings bridge the gap between reasoning capability and inference overhead, offering a practical solution for knowledge-intensive applications.
\section*{Limitations}
In this paper, we conduct the first systematic investigation into analyzing the reasoning patterns of LLMs through SVD, leveraging these findings to motivate an enhanced RAG framework. However, given that reasoning pattern characterization requires broader empirical support beyond current validation scope, our experimental analysis currently focuses on: (a) RAG system verification through external knowledge grounding, and (b) internal answer coherence validation within the LLM's parametric knowledge base. 
\section*{Acknowledgments}
This work is primarily supported by the Key Research and Development Program of China from MOST(2024YFB3311901).
We would like to express our gratitude to the anonymous ARR February reviewers (Reviewer 27qA, xzJs, VoyJ) and Area Chair TVjr for their valuable feedback and suggestions that helped improve this paper.

\bibliography{custom}

\appendix

\section{Logit Lens}
\label{sec:appendix:logit-lens}
LogitLens is an interpretability framework for analyzing language model hidden states by examining logits (raw prediction scores) and their corresponding probability distributions. Specifically, for a given hidden state $\vh_l$ at the $\vl^{th}$ layer, the logits $\vs_l$ and probabilities $\vp_l$ over the output vocabulary set $V$ are defined as:

\begin{equation}
\begin{cases}
\vs_l=\mW_U\vh_l^i\in\mathbb{R}^{|V|}, \\
\vp_l=\mathrm{softmax}\left(\vs_l\right) & 
\end{cases},
\end{equation}
where $\mW_U$ denotes the unembedding matrix, which is the same matrix used in the final layer of the model for prediction.

Our evaluation spans diverse LLM architectures, including \texttt{GPT-2-XL}, \texttt{Llama-2-7B}, and \texttt{Llama-2-13B}. We manually annotated intermediate answers for the 2WikiMultiHopQA dataset to establish ground truth for multi-hop reasoning steps.

\section{Transformation Divergence in LLMs of Varying Scales}
\label{sec:appendix:transformation-divergence}
\begin{figure}
    \centering
    \includegraphics[width=\linewidth]{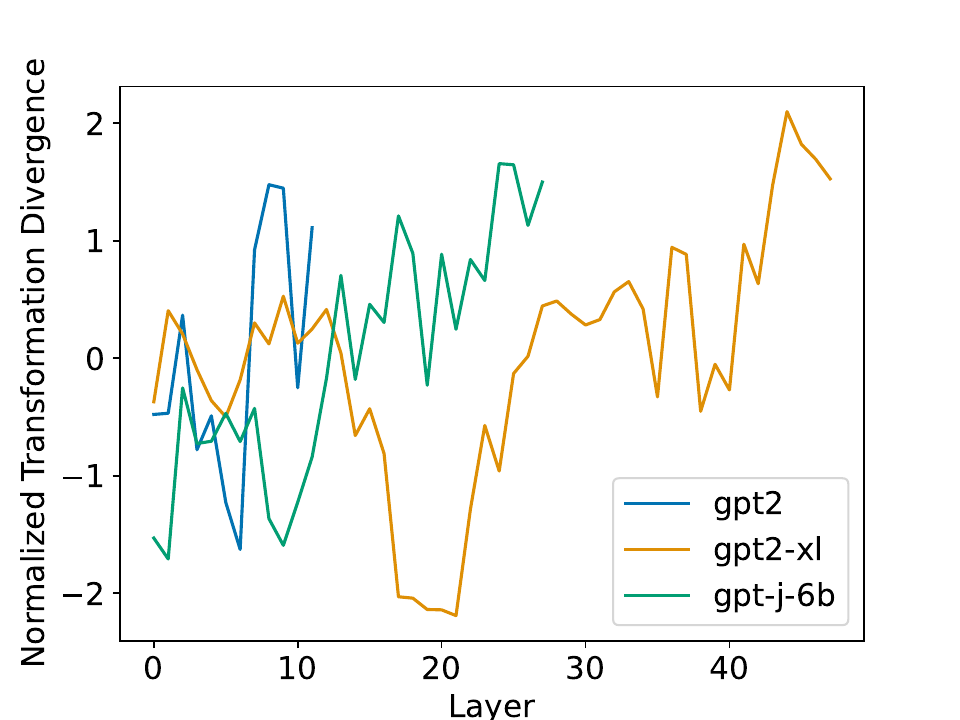}
    \caption{Transformation Divergence in gpt of Varying Scales}
    \label{fig:appendix:td-gpt}
\end{figure}

\begin{figure}
    \centering
    \includegraphics[width=\linewidth]{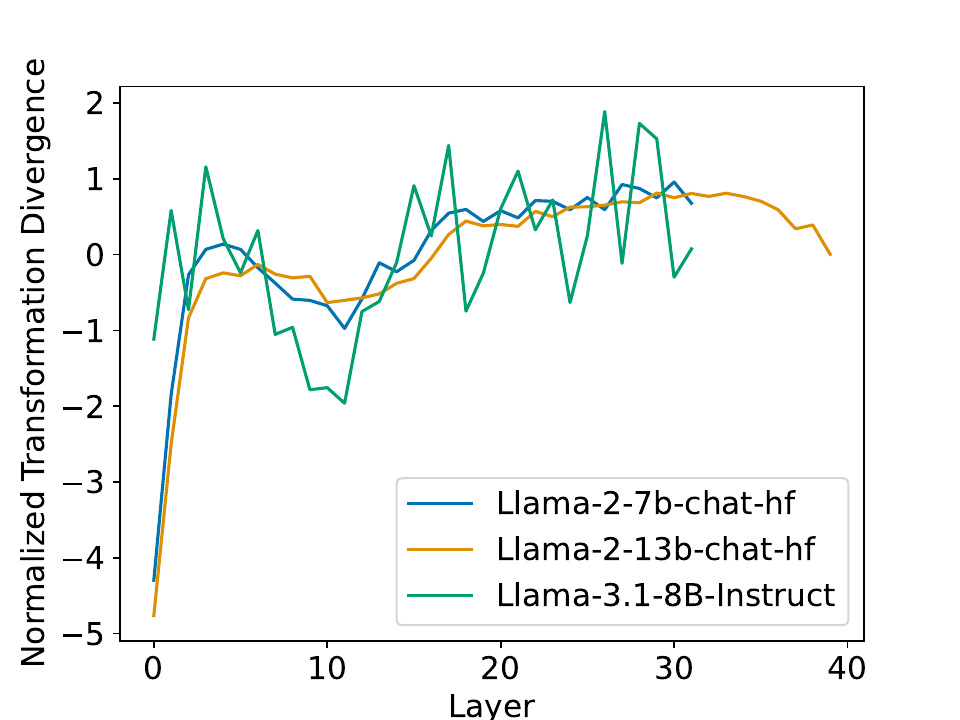}
    \caption{Transformation Divergence in Llama of Varying Scales}
    \label{fig:appendix:td-llama}
\end{figure}

We compute the transformation divergence for various scaling \texttt{GPT} models, including \texttt{GPT-2} (124M), \texttt{GPT-2-XL} (1.5B), and \texttt{GPT-J-6B}, as illustrated in Figure \ref{fig:appendix:td-gpt}, as well as for \texttt{Llama-2-7B-chat-hf}, \texttt{Llama-2-13B-chat-hf}, and \texttt{Llama-3.1-8B-Instruct}, as shown in Figure \ref{fig:appendix:td-llama}.


\section{Multi-Hop Dataset Example}
\label{sec:appendix:multi-hop-dataset-example}
The manually annotated dataset is derived from 2WikiMultiHopQA, with examples including intermediate answers, Here is an example:
\begin{description}
  \item[Question] \texttt{What is the date of birth of Mina Gerhardsen's father?} 
  \item[Document] 
  \begin{enumerate}
      \item \texttt{Mina Gerhardsen (born 14 September 1975) is a Norwegian politician for the Labour Party. She is the daughter of Rune Gerhardsen and Tove Strand, and granddaughter of Einar Gerhardsen. She is married to Eirik Øwre Thorshaug. She led the Oslo branch of Natur og Ungdom from 1993 to 1995, and was deputy leader of the Workers' Youth League in Oslo in 1997. She took the cand.mag. degree at the University of Oslo in 1998, and also has master's degrees in pedagogy from 2000 and human geography from 2003. From 1999 to 2002 she worked part-time as a journalist in \"Dagsavisen\" and \"Dagbladet\". She then worked in the Norwegian Red Cross from 2002 to 2004, except for a period from 2003 to 2004 as a journalist in \"Mandag Morgen\". She was hired as a political advisor in the Norwegian Office of the Prime Minister in 2005, when Stoltenberg's Second Cabinet assumed office. In 2009 she was promoted to State Secretary. In 2011 she changed to the Ministry of Culture.}
      \item \texttt{Rune Gerhardsen (born 13 June 1946) is a Norwegian politician, representing the Norwegian Labour Party. He is a son of Werna and Einar Gerhardsen, and attended Oslo Cathedral School. He chaired the Workers' Youth League from 1973 to 1975 and chaired the City Government of Oslo from 1992 to 1997. He chaired the Norwegian Skating Association from 1986 to 1990 and 2001 to 2003 and also 2013 to 2017.}
  \end{enumerate} 
  \item[Intermediate Answer] \texttt{Rune Gerhardsen}
  \item[Final Answer] \texttt{13 June 1946} 
\end{description}

\section{Transformation Divergence and Recall Rate}
\label{sec:td-recall-rate}
\begin{figure}
    \centering
    \includegraphics[width=\columnwidth]{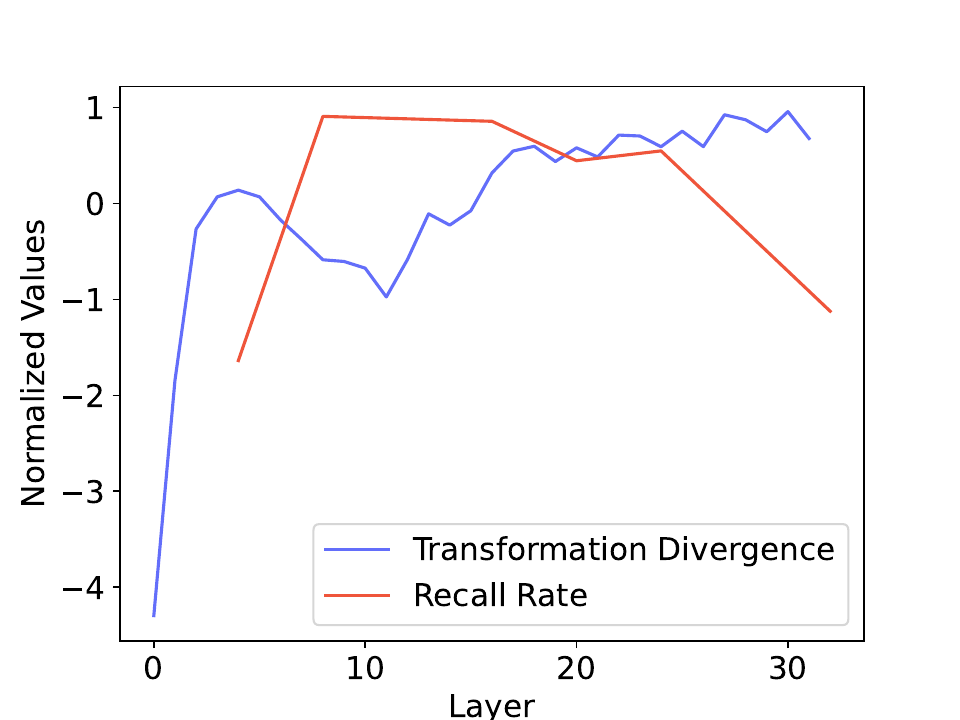}
    \caption{Transformation Divergence across the llama2-7b model and the corresponding representation generated for retriever performance.}
    \label{fig:td-recall-rate}
\end{figure}

To evaluate retriever performance with different layer representations, we utilize \texttt{meta-llama/Llama-2-7b-chat-hf} to generate representations in layers 4, 8, 16, 20, 24, 32 and compute the recall rate for the documents indirectly related to the query in 2WikiMultiHopQA dataset which result is shown in Figure \ref{fig:td-recall-rate}.

\end{document}